\documentclass[letterpaper, 10pt, twocolumn]{article}
\usepackage{style}
\usepackage{times}
\usepackage{helvet}
\usepackage{graphicx}
\usepackage{amssymb}
\usepackage{url, hyperref}
\usepackage[square,numbers]{natbib}
\begin{document}

\title{Investigating Emotion-Color Association in Deep Neural Networks}

\author{
  \textbf{Shivi Gupta}\\
  Department of Electrical Engineering\\
  Indian Institute of Technology Kanpur\\
  \texttt{shivigup@iitk.ac.in} \\
  
  \AND
  
  Shashi Kant Gupta\\
  Department of Electrical Engineering\\
  Indian Institute of Technology Kanpur\\
  \texttt{shashikg@iitk.ac.in} \\
}

\maketitle
\thispagestyle{empty}

\begin{abstract}
It has been found that representations learned by Deep Neural Networks (DNNs) correlate very well to neural responses measured in primates' brains and psychological representations exhibited by human similarity judgment. On another hand, past studies have shown that particular colors can be associated with specific emotion arousal in humans. Do deep neural networks also learn this behavior? In this study, we investigate if DNNs can learn implicit associations in stimuli, particularly, an emotion-color association between image stimuli. Our study was conducted in two parts. First, we collected human responses on a forced-choice decision task in which subjects were asked to select a color for a specified emotion-inducing image. Next, we modeled this decision task on neural networks using the similarity between deep representation (extracted using DNNs trained on object classification tasks) of the images and images of colors used in the task. We found that our model showed a fuzzy linear relationship between the two decision probabilities. This results in two interesting findings, 1. The representations learned by deep neural networks can indeed show an emotion-color association 2. The emotion-color association is not just random but involves some cognitive phenomena. Finally, we also show that this method can help us in the emotion classification task, specifically when there are very few examples to train the model. This analysis can be relevant to psychologists studying emotion-color associations and artificial intelligence researchers modeling emotional intelligence in machines or studying representations learned by deep neural networks.
\end{abstract}

\section{Introduction}
Deep learning has become interestingly popular in the machine learning community \cite{LeCun2015}. It dominates the field of computer vision, attaining near or above human-level performance in most vision tasks \cite{NIPS2012_4824, F-RCNN, kaiming2015, gans}, and has made a significant impact in natural language processing \cite{Radford2018ImprovingLU, Devlin_2019, brown2020language} and in decision-making \& control tasks in the form of deep reinforcement learning \cite{Mnih2015, Silver2016}. There's also been a tremendous increase in research on understanding the intuitive representations learned by DNNs. Looking at these state of the art performances and the structural similarity between DNNs and Biological Neural Networks, cognitive scientists have started looking at the representations learned by these models and finding whether DNNs can model human behavior. Findings show that DNNs are capable of capturing neural responses in visual tasks \cite{KhalighRazavi2014, Yamins2014, Schrimpf2018} and language comprehension \cite{NIPS2018_7897, Schrimpf2020}. Some findings also provide evidence that DNNs are also capable of capturing psychological representations. \cite{lake2015, Peterson_2017}

While deep neural networks show similarities with human representations, one fascinating question remains, can they learn implicit stimuli associations? And, can these deep neural networks show some emotional capabilities, like in humans? In this study, we try to answer this by analyzing the emotion-color association. Emotion is one of the most exciting aspects in human and is very extensively researched in emotion psychology. Different stimuli happen to elicit different kinds of emotions in humans. Psychologists have also extensively studied color perception for their special relationship with emotions, and findings suggest that different colors also elicit different emotions \cite{Munsell, Gilbert2016Sep}. Some studies have suggested that emotion-arousal is related to the visual cortex \cite{Lang1998, Krageleaaw4358}. Therefore, we decided to study this emotion-color association using deep neural networks trained on a visual task. Understanding the capability of deep neural networks to learn some emotion-color association will potentially benefit both cognitive scientists and artificial intelligence researchers. Cognitive scientists can use this to study how emotions are represented in humans and artificial intelligence researchers can use it in modeling emotional intelligence in machines. On the other hand, artificial intelligence researchers can potentially use associative stimuli as a control vector to train deep learning models for classification tasks (which we briefly explored at the end of the paper).

% In this study, we devised a method to study whether pre-trained deep neural networks, trained on object classification tasks, show an emotion-color association. 
First, we conducted a behavioral experiment of a forced-choice decision task in which subjects were asked to select a specific color for a given emotion-inducing image stimuli. We estimated decision probabilities using the responses that we got from this experiment. Next, we developed a computational model for this decision task using similarities between deep representation (extracted using DNNs trained on object classification tasks) of the stimuli images and images of colors. We then examined the relationship between the two decision probabilities using Pearson’s correlation coefficient (R). We found that the representation learned by deep neural networks indeed captures some emotion-color association. The representation from the 'fc2' layer of VGG16 showed a fuzzy linear relationship. Though this indicates the possibility of a correlation between the model and humans, it did not help much on few-shot emotion classification tasks as compared to standard classification methods. In a recent study by \cite{Peterson_2017}, the authors showed that a linearly transformed representation of the 'fc2' layer in VGG16 could represent human psychological representations. So, we tested our model after linearly transforming the raw representations. We found that the correlation score significantly improved, and the model showed a significant improvement in the emotion classification task compared to the standard classification model.

\begin{figure}[t!]
    \centering
    \includegraphics[width=0.98\linewidth]{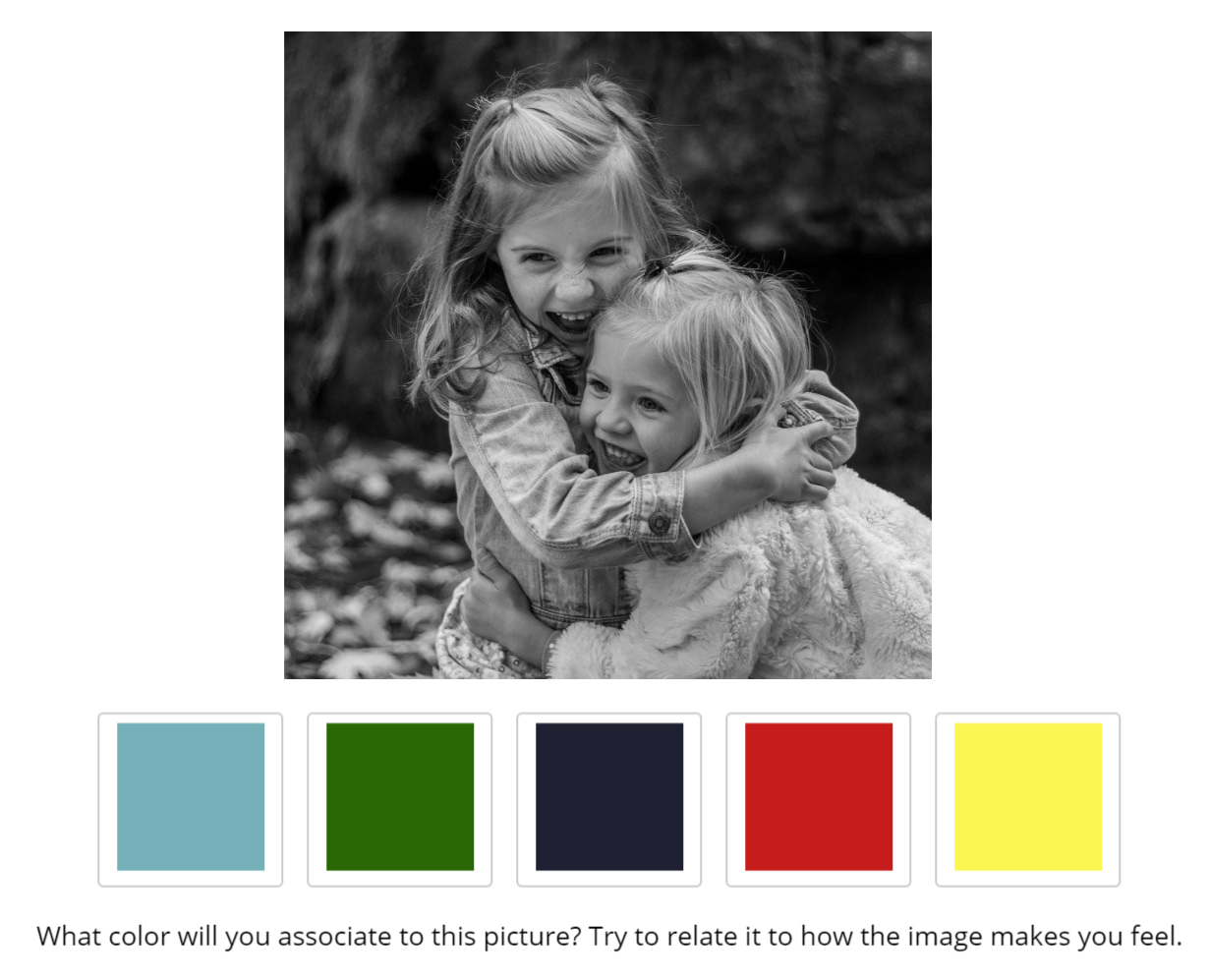}
    \caption{An illustration of the trial from the behavioral experiment. Subjects were  asked  to  select  a  single  color from the five available options. Note that the stimuli image shown here is for illustration purpose which is free to use.}
    \label{fig:exp_trial}
\end{figure}

\begin{figure*}[t]
    \centering
    \includegraphics[width=0.8\linewidth]{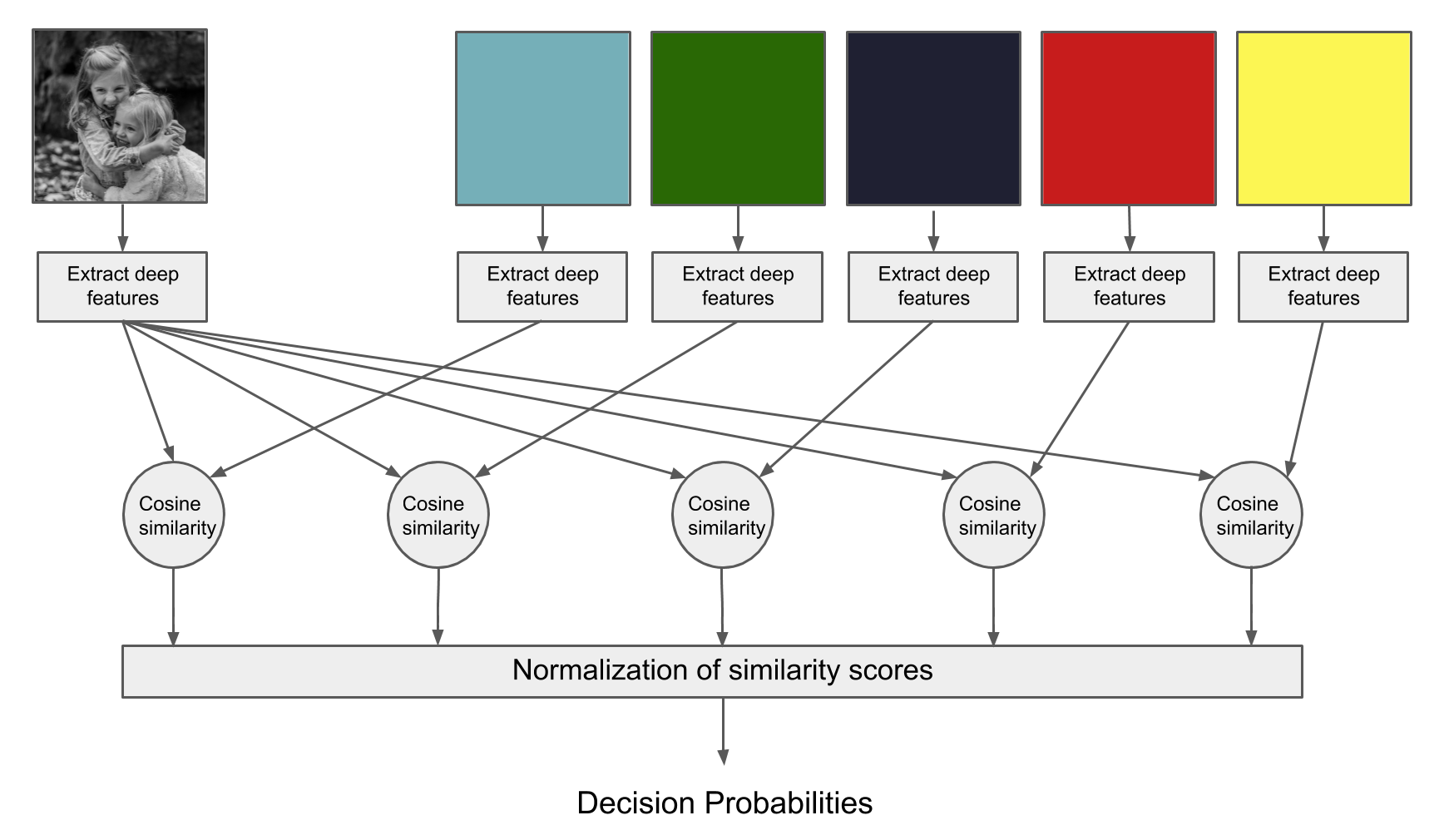}
    \caption{Modelling Decisions from Deep Representations}
    \label{fig:dnn-decision}
\end{figure*}

\section{Behavioral Experiment}\label{sec: exp}
Studying emotion-color association can have two approaches: finding colors associated with specific emotions or finding emotions associated with specific colors. In the former, there can further be two experimental approaches:
\begin{itemize}
\item Constrained (Having a small subset of colors) and unconstrained (all available colors) experiments
\item Objective (a visual color) and subjective experiments (a word for a color)
\end{itemize}

Further, in objective experiments, colors can be continuous, for example, selected using sliders or discrete, for example, chosen out of a palette. This study's experiment involves finding colors associated with emotions, and is constrained, objective, and discrete. There are six basic emotions for which color is considered as a perceptual feature: Anger (red), disgust (green), fear (black), happiness (yellow), sadness (blue), and surprise (bright), which are the colors used in other studies on emotion-color association \cite{Munsell, Gilbert2016Sep, Sutton2016Jun}. For our final version of the experiment, we only used the first five emotions as it was ambiguous to use any specific color for "bright". 

\subsection{Stimuli} 
Our stimulus set consisted of 50 grey-scaled images. These images were taken from an emotion data-set used by \cite{emotionDataset} for affective image classification. The images were selected to include 10 images for each emotion. We converted the images to gray-scale so to remove any bias because of dominant colors in the images themselves.

\subsection{Participants}
We distributed the experiment among students of the institute where this study was conducted. Their participation was completely voluntary, and none of them were forced to take part. A total of 56 different individuals completed the experiment.

\subsection{Data analysis and procedure}
\label{subsec:exp-ana}
The experiment was designed using jsPsych JavaScript library \cite{deLeeuw2015}. For an individual trial, a gray-scale image was shown along with the five colors, as mentioned before (See Figure \ref{fig:exp_trial}). At the beginning of the experiment, participants were instructed to select a color that would best fit with the underlying emotion of the shown picture. To make sure that there was no bias, we added an additional instruction to each trial, "What color will you associate to this picture? Try to relate it to how the image makes you feel". After collecting the responses, we calculated histograms of chosen colors for each image stimuli. The normalised histogram was taken as the decision probabilities of choosing colors for an image.

\section{Modelling Decisions from Deep Representations} \label{sec: dnn-decision}
A deep neural network trained on an image classification task can not only output a classification probability, but can also be used to extract different feature representations of an input image by extracting outputs from intermediate layers in the network. We extracted these features using state of the art deep learning models trained on the imagenet dataset \cite{imagenet}. We refer to these extracted features as deep representations. We used these deep representations to model decision probabilities of selecting colors for a given stimuli image. The process is explained as follows: 

\begin{itemize}
\item Firstly, we extracted deep representations of stimuli images and color images by passing them through a pre-trained deep learning model (Input images were resized to 224 x 224 pixels). We used VGG16, DenseNet, ResNet, and MobileNet architectures for our study \cite{vgg16, densenet, resnet, mobilenets}.
\item Then we calculated cosine similarities between extracted representations of stimuli images and color images.
\item Finally, to get overall decision probabilities from our model, we normalised the similarity scores for a given stimuli image among the five color images.
\end{itemize}

So, our decision model outputs a probability of choosing a particular color for a given image. We try to see if these predictions are similar to the data collected using the behavioral experiment (Refer to next Section).

\section{Evaluating Correspondence} \label{sec:eval-rep}
To evaluate the correspondence between pre-trained deep neural network models and human decisions, we followed a method introduced by \cite{Peterson_2017} for comparing psychological representations with deep representations extracted using DNNs. Their study evaluated the correspondence for similarity judgements by calculating correlation between experimental data and similarity generated using deep representations. The decision probabilities in our study are modelled using the similarity between the representations of stimuli images and color images. So, evaluating the correspondence between the two decision probabilities is similar to evaluating the correspondence between the similarity judgement of humans and DNNs. To evaluate the correspondence, we find the Pearson's correlation coefficient (R) between the two decision probabilities.

\subsection{Extracting Features}
We used VGG16, DenseNet169, ResNet, and MobileNet as our DNN models to extract deep representations. We used the pre-trained weights provided by TensorFlow deep learning library. All the models were trained on the imagenet dataset to classify 1000 object categories. For our analysis, we mostly used the last layer of each model before the final classification layer to extract image features because the analysis from \cite{Peterson_2017} shows that deeper layers are better at predicting psychological representations. Therefore, the number of features extracted using VGG16 is 4096, DenseNet169 is 1664, ResNet50 is 2048, and MobileNetv2 is 1000.

\begin{table}[t!]
    \centering
    \begin{tabular}{|c|c|c|c|c|}
        \hline
        Model & Layer & $R_{r}$ & $R_{t}$\\
        \hline
        VGG16 & fc2 & \textbf{$0.33$} & \textbf{$0.63 \pm 0.01$}\\
        \hline
        DenseNet169 & avg\_pool & $0.29$ & $0.57 \pm 0.03$\\
        \hline
        ResNet50 & avg\_pool & $0.28$ & $0.60 \pm 0.02$\\
        \hline
        MobileNet & reshape\_2 & $0.26$ & $0.53 \pm 0.03$\\
        \hline
    \end{tabular} 
    \caption{R scores found using various pre-trained deep learning models. Layer names are as per the names in the models in TensorFlow deep learning library. The third column (i.e. $R_{r}$) corresponds to the R score calculated using raw representations and the fourth column (i.e. $R_{t}$) corresponds to the R score calculated using transformed representations. For transformed representation, mean scores and standard deviations are reported over 50 independent runs.}
    \label{tab:adv_table}
\end{table} 

\subsection{Results}
Results are shown in Table \ref{tab:adv_table} (Column 3, i.e., $R_{r}$ is the correlation score evaluated using raw representation. $R_{t}$ is the correlation score evaluated using transformed representations, explained in next section). VGG16 showed the best results with a correlation score of $R = 0.33$ with $pvalue < 0.0001$ (null hypothesis being zero correlation). According to statistics, a correlation score of $0.3 < R < 0.7$ represents moderate linear relationship (fuzzy linear relationship) \cite{Ratner2009}. While $R = 0.33$ indicates a moderate linear relationship between the model decisions and human decisions, the score is still small. So, before making any claims, we checked the $R$ scores against wrong colored images, i.e., we changed labels of the colored images, so as to result in wrong similarity scores for image-color pairs. For example, if we initially had labels 0, 1, 2, 3, and 4 for colors red, green, blue, black, and yellow respectively, we changed these labels to 4, 3, 0, 2, and 1. So our decision model will assume the deep representations of yellow color (label 4) to be the deep representations for red color (label 0); this will similarly change for the other pairs. So this implies that we will be comparing the human decision probability of associating a stimuli image with yellow color with the model's decision probability for associating the same stimuli image with red color. We found that this decreases the R score significantly (See Table \ref{tab:Rscore}, Column 2). This indicates that images associate with specific colors. If $R = 0.33$ wasn't representing any emotion-color association, we would have obtained comparable $R$ scores after changing color labels. But that is not the case. This is an interesting result for cognitive scientists to explore emotion-color associations learned in humans. This may still not be of great advantage to AI researchers, so we further worked on improving this emotion-color association for DNNs. The low $R$ score could be attributed to the following reasons: 1. There's no straight association between color and emotion-inducing images or 2. The features extracted using VGG16 don't directly correspond to representations of emotions and need to be transformed to some other dimension, which could better associate with color and emotions. We tested for the second cause, similar to the method proposed by \cite{Peterson_2017} and \cite{jha2020extracting} on capturing human similarity using deep neural networks.

\begin{table}[t!]
    \centering
    \begin{tabular}{|c|c|c|}
        \hline
        Color sequence & R & R (transformed)\\
        \hline
        [0, 1, 2, 3, 4] (original seq.) & 0.33 & 0.63 $\pm$ 0.01\\
        \hline
        [4, 3, 0, 2, 1] & 0.01 & 0.04\\
        \hline
        [2, 3, 1, 4, 0] & -0.09 & -0.34\\
        \hline
        [2, 4, 3, 1, 0] & 0.11 & 0.05\\
        \hline
        [1, 0, 4, 2, 3] & 0.06 & -0.03\\
        \hline
    \end{tabular} 
    \caption{R score against wrong color labels. The second column contains the R score calculated using raw representations and the third column contains the R score from transformed representations. The first row corresponds to original labels of the colors. Except the original sequence, other sequence were evaluated for a single iteration.}
    \label{tab:Rscore}
\end{table} 

\begin{figure*}[t]
    \centering
    \includegraphics[width=0.8\linewidth]{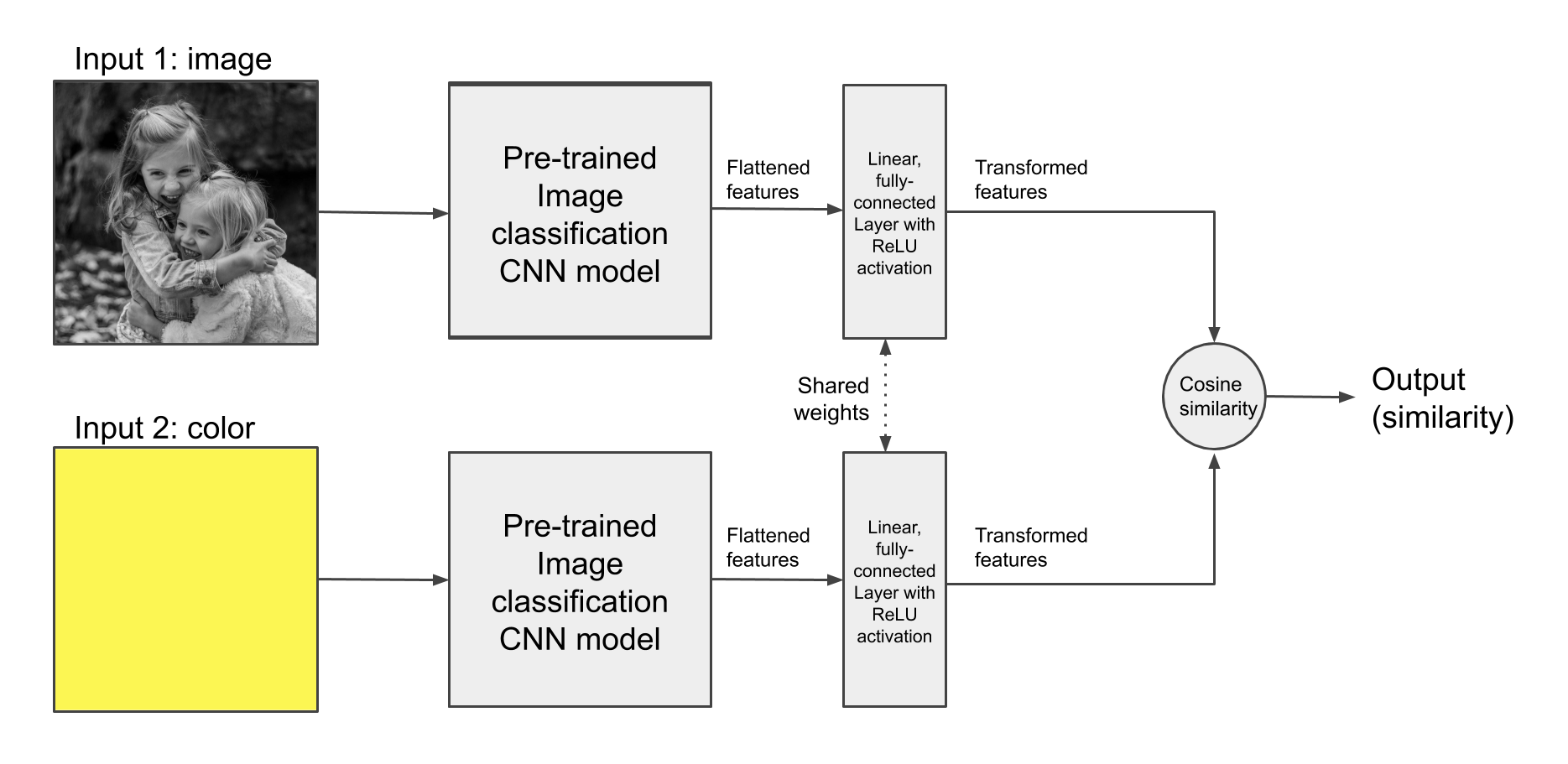}
    \caption{Feature Transformation}
    \label{fig:transformed}
\end{figure*}

\section{Evaluating Transformed Representation} \label{sec: eval_trans_rep}
In the last section, we showed that deep neural networks indeed capture some aspect of emotion-color relation but the correlation calculated was still small. So, we evaluated the possibility of achieving a higher correlation score. \cite{Peterson_2017} and \cite{jha2020extracting} showed a method to transform deep representations to capture psychological representations of human similarity judgement. In their studies, they extracted features using a pre-trained deep learning model and then linearly transformed the features to a higher dimensional feature space such that the similarity score between the two images can better predict human similarity judgement on the same pair of images. On similar lines, we first found a linear transform of the deep representations extracted using pre-trained models to a smaller number of features. Note that our cosine similarity should range between 0 and 1; therefore, we clip the output values at zero by applying a ReLU activation. And then, we performed the previous analysis as we did for raw representations on the transformed representations. The process is illustrated in figure \ref{fig:transformed}. We first tested for various number of output features starting from 0 to 175 with a step size of 25. We found that the correlation score maximized around output features = 75. So, we used 75 numbers of output features for further analysis. We trained the weights for this linear layer using the similarity scores obtained from the behavioral experiment. We used L2 loss function between the human similarity and similarity predicted by the model. The model was evaluated using five fold cross-validation for its generalisation performance. Note that we have a total of 250 different similarity scores corresponding to 50 different stimuli images and 5 color images. For each cross-validation set, only 200 similarity pairs were used for training, and rest 50 were used for model evaluation. Our training parameters are: adam optimiser with learning rate = 0.001, batch size = 10, and number of epochs = 30. During training, we shuffled the 200 input data.

\subsection{Results}
Results are shown in Table \ref{tab:adv_table} (Column 4, i.e. $R_{t}$). The R scores reported here are calculated using the similarity obtained on the validation set of the five fold cross-validation method. So, for each fold, we get 50 similarity scores corresponding to the validation set of that fold, comprising of a total of 250 similarities for the overall run. Also, note that reported results are averaged over 50 independent runs (we have reported mean along with standard deviation). We found that the R score significantly improved for the transformed representation for all the four models. Most importantly, VGG16 performed best (with $R = 0.63$ and $pvalue < 0.0001$ ) which is consistent with the evaluation done in \cite{Peterson_2017} on psychological representation (they also observed that VGG16 produces best result on capturing psychological representations). We also performed the analysis on wrong classes for transformed representation on VGG16 to confirm if this transformation is not arising out of large number of features extracted from VGG16. Interestingly, we found that the R-scores for wrong color labels were significantly low than the correct color labels (Table \ref{tab:Rscore}, Column 3). 
\begin{figure}[b!]
    \centering
    \includegraphics[width=0.98\linewidth]{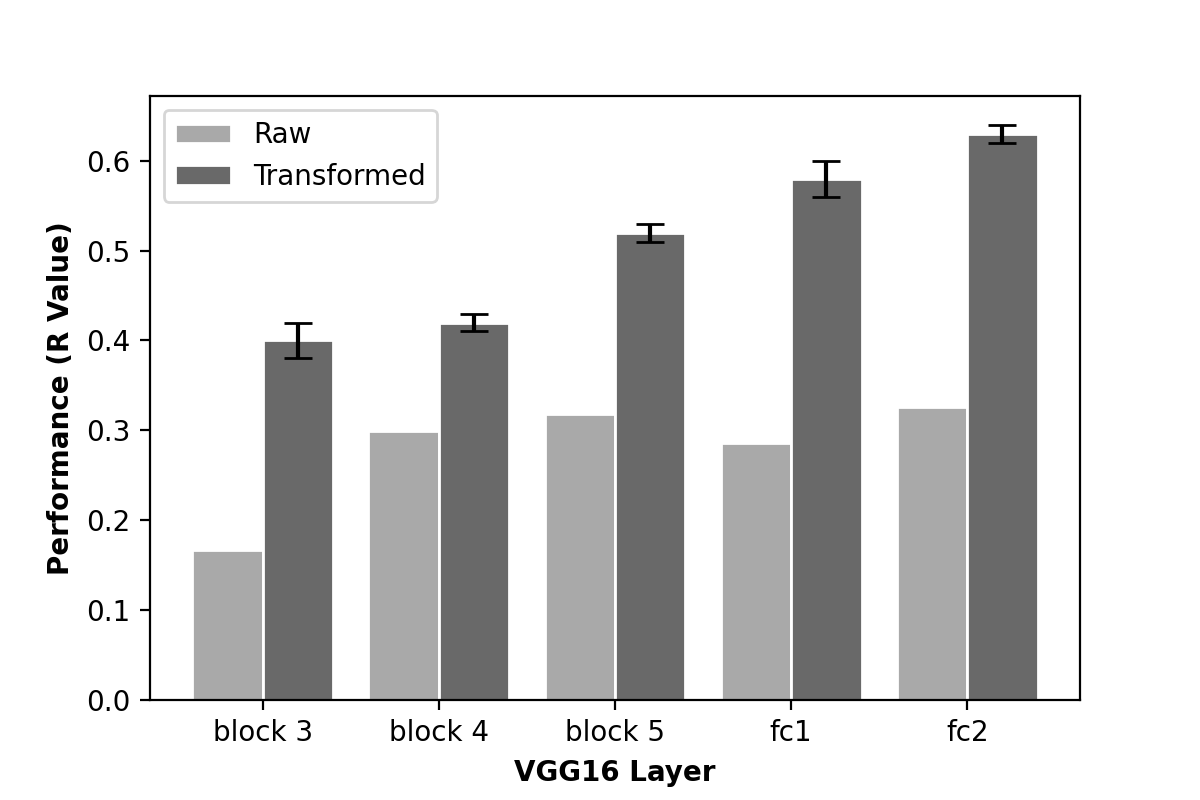}
    \caption{VGG 16 performance across different pooling layers. Bars shows the average accuracy over 10 trials and the error bars show the standard deviation. For 'fc2' accuracy is averaged over 50 trials.}
    \label{fig:bars_vgg16}
\end{figure}
We also evaluated features extracted across different pooling layers of VGG16 to check if they produce similar trends as with psychological representation \cite{Peterson_2017} i.e., deeper layers better capture human judgement. Figure \ref{fig:bars_vgg16} shows the obtained result, bars show the average accuracy over 10 trials, and the error bars show the standard deviation. We found that the results are indeed valid with the results of \cite{Peterson_2017} on psychological representations.

\section{Evaluating on Classification Task}
After studying the possibility of deep neural networks to predict emotion-color association, we evaluated the method by its ability to predict emotions for a given emotion-inducing image. The predicted emotion will simply be the emotion corresponding to the color with which it showed maximum similarity. We considered two cases. In case 1, we took true class label to be the class label predicted by humans in our experimental study (class for a given stimulus was taken to be the color label which got maximum votes). In case 2, we took true class label to be the class label provided in the original dataset. We do this to compare different methods better, considering there isn't a 100\% correlation between colors and emotions. We compared the following four methods:

\begin{itemize}
    \item Raw similarity: No training involved; we predict the classes based on the color which gives maximum similarity to the input images based on the features extracted using the 'fc2' layer of VGG16.
    \item Transformed similarity: We predict the classes based on the color which gives maximum similarity to the input images based on the transformed representation. We trained the model using five fold cross-validation, and the accuracy reported here is based on the label predicted using test cases 'only'. The training parameters were the same, which we used previously.
    \item Standard classification (on human prediction): We replaced the last layer of VGG16 with a fully connected layer with 75 output units and 'relu' activation and then added one another layer to give five outputs and 'softmax' function to predict among 5 class labels. We trained the model on human prediction using categorical cross-entropy loss. The reported accuracy is for test cases only in a five fold cross-validation. Our training parameters were: adam optimiser with learning rate = 0.001, batch size = 10, and number of epochs = 15. (We trained it using fewer epochs compared to the similarity model, because this model converges faster than the similarity model. Even if we take epochs = 30, the results were not significantly different).
    \item Standard classification (on actual classes): Similar to the previous one, but the model was trained using actual class labels.
\end{itemize}

\subsection{Results}
Table \ref{tab:classification} shows results for the classification tasks. Accuracy reported is average accuracy over 50 training trials. We were amazed to find that emotion classification using raw representations yields 40\% classification accuracy for case 1, which is way above chance (~8\% accuracy). It's also important to note that we did not explicitly train our model to classify to those specific emotions that humans predicted. This further validates our point that Deep Neural Networks are capable of capturing emotion-color association. The model's performance further increased after we linearly transformed the features, achieving 40.20\% on actual classes and 56.12\% on human predictions. In both of the cases, the similarity-based model performed better than the standard classification model. We also compared the maximum accuracy achieved by different models among the 50 trials. For the similarity based model (transformed representation), max accuracy achieved was 46\% on actual class and 64\% on human prediction. While the Standard classification model (trained on human prediction) achieved 40\% accuracy on actual class and 52\% accuracy on human prediction. Considering the accuracy for human prediction to actual class (~54\%), the data size of just 50 images and difficulty of the task, the amount of generality shown by the model is remarkably well and is clearly way better than a chance performance.

\begin{table}[t!]
    \centering
    \begin{tabular}{|c|c|c|}
        \hline
        Method & wrt Human & wrt Actual\\
         & Prediction & Class\\
        \hline
        Chance (1000 runs) & 7.9 $\pm$ 2.88 & 20.01 $\pm$ 4.44\\
        \hline
        Raw similarity & 40 & 24\\
        \hline
        Transformed similarity & \textbf{56.12 $\pm$ 3.21 }& \textbf{40.20 $\pm$ 2.89}\\
        \hline
        Standard classification &  &  \\
        (on human prediction) & 43.84 $\pm$ 4.70 & 32.60 $\pm$ 3.49\\
        \hline
        Standard classification &  &  \\
        (on actual class) & 31.80 $\pm$ 5.49 & 30.44 $\pm$ 4.74\\
        \hline
    \end{tabular} 
    \caption{Accuracy with respect to human prediction and actual class labels.}
    \label{tab:classification}
\end{table} 

\section{Discussion}
In this analysis, we show that representations learned by Deep Neural Networks are capable of capturing emotion-color association. Though comparing raw representations yielded a low correlation score, the representations show a greater generality and correlation to human decisions when linearly transformed. We also showed how we could use this overall method to train deep learning models for an emotion classification task.  Our analysis answers an interesting question in Cognitive Sciences. The human emotion-color association is not random but is learned while performing other cognitive tasks. If not, wrong labeled colors would have shown higher correlation scores for the transformed representation, as the network was exclusively trained to do that. But we see a big difference between the correlation scores of correctly labeled colors and wrong labeled ones. The method could also be very beneficial to the Machine Learning community on finding alternative ways to train deep learning models for classification problems, which could probably improve the performance when we have smaller datasets. However, a potential limitation of this would be that you will need to identify which alternate associative feature to use for a specific task; for example, we used colors for emotion classification. The study also needs more and more replication work on different datasets to validate the point for the generality of this method to study stimuli association in deep neural networks.

We see great potential for this result and method for advancement in affective computing in developing artificial emotional intelligence and in emotion psychology. And we envision that both the community will expand these results to greater heights.

\section*{Contributions}

The study was designed by Shashi Kant Gupta (S.K.G.). Shivi Gupta (S.G.) performed the experiment, data analyses, and modelling under the technical guidance and supervision of S.K.G. S.G. and S.K.G. wrote the manuscript.

\begin{small}
    \bibliography{main}

\begin{thebibliography}{31}
\providecommand{\natexlab}[1]{#1}
\providecommand{\url}[1]{\texttt{#1}}
\expandafter\ifx\csname urlstyle\endcsname\relax
  \providecommand{\doi}[1]{doi: #1}\else
  \providecommand{\doi}{doi: \begingroup \urlstyle{rm}\Url}\fi

\bibitem[Brown et~al.(2020)Brown, Mann, Ryder, Subbiah, Kaplan, Dhariwal,
  Neelakantan, Shyam, Sastry, Askell, Agarwal, Herbert-Voss, Krueger, Henighan,
  Child, Ramesh, Ziegler, Wu, Winter, Hesse, Chen, Sigler, Litwin, Gray, Chess,
  Clark, Berner, McCandlish, Radford, Sutskever, and Amodei]{brown2020language}
T.~B. Brown, B.~Mann, N.~Ryder, M.~Subbiah, J.~Kaplan, P.~Dhariwal,
  A.~Neelakantan, P.~Shyam, G.~Sastry, A.~Askell, S.~Agarwal, A.~Herbert-Voss,
  G.~Krueger, T.~Henighan, R.~Child, A.~Ramesh, D.~M. Ziegler, J.~Wu,
  C.~Winter, C.~Hesse, M.~Chen, E.~Sigler, M.~Litwin, S.~Gray, B.~Chess,
  J.~Clark, C.~Berner, S.~McCandlish, A.~Radford, I.~Sutskever, and D.~Amodei.
\newblock Language models are few-shot learners, 2020.

\bibitem[D'Andrade and EGAN(2009)]{Munsell}
R.~D'Andrade and M.~EGAN.
\newblock The colors of emotion.
\newblock \emph{American Ethnologist}, 1:\penalty0 49 -- 63, 10 2009.
\newblock \doi{10.1525/ae.1974.1.1.02a00030}.

\bibitem[de~Leeuw and Motz(2015)]{deLeeuw2015}
J.~R. de~Leeuw and B.~A. Motz.
\newblock Psychophysics in a web browser? comparing response times collected
  with {JavaScript} and psychophysics toolbox in a visual search task.
\newblock \emph{Behavior Research Methods}, 48\penalty0 (1):\penalty0 1--12,
  Mar. 2015.
\newblock \doi{10.3758/s13428-015-0567-2}.
\newblock URL \url{https://doi.org/10.3758/s13428-015-0567-2}.

\bibitem[Deng et~al.(2009)Deng, Dong, Socher, Li, Li, and Fei-Fei]{imagenet}
J.~Deng, W.~Dong, R.~Socher, L.-J. Li, K.~Li, and L.~Fei-Fei.
\newblock {ImageNet: A Large-Scale Hierarchical Image Database}.
\newblock In \emph{CVPR09}, 2009.

\bibitem[Devlin et~al.(2019)Devlin, Chang, Lee, and Toutanova]{Devlin_2019}
J.~Devlin, M.-W. Chang, K.~Lee, and K.~Toutanova.
\newblock {BERT}: Pre-training of deep bidirectional transformers for language
  understanding.
\newblock In \emph{Proceedings of the 2019 Conference of the North {A}merican
  Chapter of the Association for Computational Linguistics: Human Language
  Technologies, Volume 1 (Long and Short Papers)}, pages 4171--4186,
  Minneapolis, Minnesota, June 2019. Association for Computational Linguistics.
\newblock \doi{10.18653/v1/N19-1423}.
\newblock URL \url{https://www.aclweb.org/anthology/N19-1423}.

\bibitem[Gilbert et~al.(2016)Gilbert, Fridlund, and Lucchina]{Gilbert2016Sep}
A.~N. Gilbert, A.~J. Fridlund, and L.~A. Lucchina.
\newblock {The color of emotion: A metric for implicit color associations}.
\newblock \emph{Food Qual. Preference}, 52:\penalty0 203--210, Sep 2016.
\newblock ISSN 0950-3293.
\newblock \doi{10.1016/j.foodqual.2016.04.007}.

\bibitem[Goodfellow et~al.(2014)Goodfellow, Pouget-Abadie, Mirza, Xu,
  Warde-Farley, Ozair, Courville, and Bengio]{gans}
I.~J. Goodfellow, J.~Pouget-Abadie, M.~Mirza, B.~Xu, D.~Warde-Farley, S.~Ozair,
  A.~Courville, and Y.~Bengio.
\newblock Generative adversarial nets.
\newblock In \emph{Proceedings of the 27th International Conference on Neural
  Information Processing Systems - Volume 2}, NIPS'14, page 2672–2680,
  Cambridge, MA, USA, 2014. MIT Press.

\bibitem[{He} et~al.(2015){He}, {Zhang}, {Ren}, and {Sun}]{kaiming2015}
K.~{He}, X.~{Zhang}, S.~{Ren}, and J.~{Sun}.
\newblock Delving deep into rectifiers: Surpassing human-level performance on
  imagenet classification.
\newblock In \emph{2015 IEEE International Conference on Computer Vision
  (ICCV)}, pages 1026--1034, 2015.

\bibitem[He et~al.(2016)He, Zhang, Ren, and Sun]{resnet}
K.~He, X.~Zhang, S.~Ren, and J.~Sun.
\newblock Deep residual learning for image recognition.
\newblock \emph{2016 IEEE Conference on Computer Vision and Pattern Recognition
  (CVPR)}, Jun 2016.
\newblock \doi{10.1109/cvpr.2016.90}.
\newblock URL \url{http://dx.doi.org/10.1109/cvpr.2016.90}.

\bibitem[Howard et~al.(2017)Howard, Zhu, Chen, Kalenichenko, Wang, Weyand,
  Andreetto, and Adam]{mobilenets}
A.~G. Howard, M.~Zhu, B.~Chen, D.~Kalenichenko, W.~Wang, T.~Weyand,
  M.~Andreetto, and H.~Adam.
\newblock Mobilenets: Efficient convolutional neural networks for mobile vision
  applications, 2017.

\bibitem[Huang et~al.(2017)Huang, Liu, Van Der~Maaten, and
  Weinberger]{densenet}
G.~Huang, Z.~Liu, L.~Van Der~Maaten, and K.~Q. Weinberger.
\newblock Densely connected convolutional networks.
\newblock \emph{2017 IEEE Conference on Computer Vision and Pattern Recognition
  (CVPR)}, Jul 2017.
\newblock \doi{10.1109/cvpr.2017.243}.
\newblock URL \url{http://dx.doi.org/10.1109/CVPR.2017.243}.

\bibitem[Jain and Huth(2018)]{NIPS2018_7897}
S.~Jain and A.~Huth.
\newblock Incorporating context into language encoding models for fmri.
\newblock In S.~Bengio, H.~Wallach, H.~Larochelle, K.~Grauman, N.~Cesa-Bianchi,
  and R.~Garnett, editors, \emph{Advances in Neural Information Processing
  Systems 31}, pages 6628--6637. Curran Associates, Inc., 2018.

\bibitem[Jha et~al.(2020)Jha, Peterson, and Griffiths]{jha2020extracting}
A.~Jha, J.~Peterson, and T.~L. Griffiths.
\newblock Extracting low-dimensional psychological representations from
  convolutional neural networks, 2020.

\bibitem[Khaligh-Razavi and Kriegeskorte(2014)]{KhalighRazavi2014}
S.-M. Khaligh-Razavi and N.~Kriegeskorte.
\newblock Deep supervised, but not unsupervised, models may explain {IT}
  cortical representation.
\newblock \emph{{PLoS} Computational Biology}, 10\penalty0 (11):\penalty0
  e1003915, Nov. 2014.
\newblock \doi{10.1371/journal.pcbi.1003915}.
\newblock URL \url{https://doi.org/10.1371/journal.pcbi.1003915}.

\bibitem[Kragel et~al.(2019)Kragel, Reddan, LaBar, and Wager]{Krageleaaw4358}
P.~A. Kragel, M.~C. Reddan, K.~S. LaBar, and T.~D. Wager.
\newblock Emotion schemas are embedded in the human visual system.
\newblock \emph{Science Advances}, 5\penalty0 (7), 2019.
\newblock \doi{10.1126/sciadv.aaw4358}.
\newblock URL \url{https://advances.sciencemag.org/content/5/7/eaaw4358}.

\bibitem[Krizhevsky et~al.(2012)Krizhevsky, Sutskever, and
  Hinton]{NIPS2012_4824}
A.~Krizhevsky, I.~Sutskever, and G.~E. Hinton.
\newblock Imagenet classification with deep convolutional neural networks.
\newblock In F.~Pereira, C.~J.~C. Burges, L.~Bottou, and K.~Q. Weinberger,
  editors, \emph{Advances in Neural Information Processing Systems 25}, pages
  1097--1105. Curran Associates, Inc., 2012.

\bibitem[Lake et~al.(2015)Lake, Zaremba, Fergus, and Gureckis]{lake2015}
B.~Lake, W.~Zaremba, R.~Fergus, and T.~Gureckis.
\newblock Deep neural networks predict category typicality ratings for images.
\newblock In R.~Dale, C.~Jennings, P.~Maglio, T.~Matlock, D.~Noelle,
  A.~Warlaumont, and J.~Yoshimi, editors, \emph{Proceedings of the 37th Annual
  Conference of the Cognitive Science Society}. Cognitive Science Society,
  2015.

\bibitem[Lang et~al.(1998)Lang, Bradley, Fitzsimmons, Cuthbert, Scott, Moulder,
  and Nangia]{Lang1998}
P.~J. Lang, M.~M. Bradley, J.~R. Fitzsimmons, B.~N. Cuthbert, J.~D. Scott,
  B.~Moulder, and V.~Nangia.
\newblock Emotional arousal and activation of the visual cortex: An {fMRI}
  analysis.
\newblock \emph{Psychophysiology}, 35\penalty0 (2):\penalty0 199--210, Mar.
  1998.
\newblock \doi{10.1111/1469-8986.3520199}.
\newblock URL \url{https://doi.org/10.1111/1469-8986.3520199}.

\bibitem[LeCun et~al.(2015)LeCun, Bengio, and Hinton]{LeCun2015}
Y.~LeCun, Y.~Bengio, and G.~Hinton.
\newblock Deep learning.
\newblock \emph{Nature}, 521\penalty0 (7553):\penalty0 436--444, May 2015.
\newblock \doi{10.1038/nature14539}.
\newblock URL \url{https://doi.org/10.1038/nature14539}.

\bibitem[Machajdik and Hanbury(2010)]{emotionDataset}
J.~Machajdik and A.~Hanbury.
\newblock Affective image classification using features inspired by psychology
  and art theory.
\newblock In \emph{Proceedings of the 18th ACM International Conference on
  Multimedia}, MM '10, page 83–92, New York, NY, USA, 2010. Association for
  Computing Machinery.
\newblock ISBN 9781605589336.
\newblock \doi{10.1145/1873951.1873965}.
\newblock URL \url{https://doi.org/10.1145/1873951.1873965}.

\bibitem[Mnih et~al.(2015)Mnih, Kavukcuoglu, Silver, Rusu, Veness, Bellemare,
  Graves, Riedmiller, Fidjeland, Ostrovski, Petersen, Beattie, Sadik,
  Antonoglou, King, Kumaran, Wierstra, Legg, and Hassabis]{Mnih2015}
V.~Mnih, K.~Kavukcuoglu, D.~Silver, A.~A. Rusu, J.~Veness, M.~G. Bellemare,
  A.~Graves, M.~Riedmiller, A.~K. Fidjeland, G.~Ostrovski, S.~Petersen,
  C.~Beattie, A.~Sadik, I.~Antonoglou, H.~King, D.~Kumaran, D.~Wierstra,
  S.~Legg, and D.~Hassabis.
\newblock Human-level control through deep reinforcement learning.
\newblock \emph{Nature}, 518\penalty0 (7540):\penalty0 529--533, Feb. 2015.
\newblock \doi{10.1038/nature14236}.
\newblock URL \url{https://doi.org/10.1038/nature14236}.

\bibitem[Peterson et~al.(2017)Peterson, Abbott, and Griffiths]{Peterson_2017}
J.~C. Peterson, J.~T. Abbott, and T.~L. Griffiths.
\newblock Adapting deep network features to capture psychological
  representations: An abridged report.
\newblock \emph{Proceedings of the Twenty-Sixth International Joint Conference
  on Artificial Intelligence}, Aug 2017.
\newblock \doi{10.24963/ijcai.2017/697}.
\newblock URL \url{http://dx.doi.org/10.24963/ijcai.2017/697}.

\bibitem[Radford(2018)]{Radford2018ImprovingLU}
A.~Radford.
\newblock Improving language understanding by generative pre-training.
\newblock 2018.

\bibitem[Ratner(2009)]{Ratner2009}
B.~Ratner.
\newblock The correlation coefficient: Its values range between +1/-1, or do
  they?
\newblock \emph{Journal of Targeting, Measurement and Analysis for Marketing},
  17\penalty0 (2):\penalty0 139--142, May 2009.
\newblock \doi{10.1057/jt.2009.5}.

\bibitem[{Ren} et~al.(2017){Ren}, {He}, {Girshick}, and {Sun}]{F-RCNN}
S.~{Ren}, K.~{He}, R.~{Girshick}, and J.~{Sun}.
\newblock Faster r-cnn: Towards real-time object detection with region proposal
  networks.
\newblock \emph{IEEE Transactions on Pattern Analysis and Machine
  Intelligence}, 39\penalty0 (6):\penalty0 1137--1149, 2017.

\bibitem[Schrimpf et~al.(2020{\natexlab{a}})Schrimpf, Blank, Tuckute, Kauf,
  Hosseini, Kanwisher, Tenenbaum, and Fedorenko]{Schrimpf2020}
M.~Schrimpf, I.~Blank, G.~Tuckute, C.~Kauf, E.~A. Hosseini, N.~Kanwisher,
  J.~Tenenbaum, and E.~Fedorenko.
\newblock The neural architecture of language: Integrative reverse-engineering
  converges on a model for predictive processing.
\newblock \emph{bioRxiv}, 2020{\natexlab{a}}.
\newblock \doi{10.1101/2020.06.26.174482}.
\newblock URL
  \url{https://www.biorxiv.org/content/early/2020/10/09/2020.06.26.174482}.

\bibitem[Schrimpf et~al.(2020{\natexlab{b}})Schrimpf, Kubilius, Hong, Majaj,
  Rajalingham, Issa, Kar, Bashivan, Prescott-Roy, Geiger, Schmidt, Yamins, and
  DiCarlo]{Schrimpf2018}
M.~Schrimpf, J.~Kubilius, H.~Hong, N.~J. Majaj, R.~Rajalingham, E.~B. Issa,
  K.~Kar, P.~Bashivan, J.~Prescott-Roy, F.~Geiger, K.~Schmidt, D.~L.~K. Yamins,
  and J.~J. DiCarlo.
\newblock Brain-score: Which artificial neural network for object recognition
  is most brain-like?
\newblock \emph{bioRxiv}, 2020{\natexlab{b}}.
\newblock \doi{10.1101/407007}.
\newblock URL \url{https://www.biorxiv.org/content/early/2020/01/02/407007}.

\bibitem[Silver et~al.(2016)Silver, Huang, Maddison, Guez, Sifre, van~den
  Driessche, Schrittwieser, Antonoglou, Panneershelvam, Lanctot, Dieleman,
  Grewe, Nham, Kalchbrenner, Sutskever, Lillicrap, Leach, Kavukcuoglu, Graepel,
  and Hassabis]{Silver2016}
D.~Silver, A.~Huang, C.~J. Maddison, A.~Guez, L.~Sifre, G.~van~den Driessche,
  J.~Schrittwieser, I.~Antonoglou, V.~Panneershelvam, M.~Lanctot, S.~Dieleman,
  D.~Grewe, J.~Nham, N.~Kalchbrenner, I.~Sutskever, T.~Lillicrap, M.~Leach,
  K.~Kavukcuoglu, T.~Graepel, and D.~Hassabis.
\newblock Mastering the game of go with deep neural networks and tree search.
\newblock \emph{Nature}, 529\penalty0 (7587):\penalty0 484--489, Jan. 2016.
\newblock \doi{10.1038/nature16961}.
\newblock URL \url{https://doi.org/10.1038/nature16961}.

\bibitem[Simonyan and Zisserman(2014)]{vgg16}
K.~Simonyan and A.~Zisserman.
\newblock Very deep convolutional networks for large-scale image recognition,
  2014.

\bibitem[Sutton and Altarriba(2016)]{Sutton2016Jun}
T.~M. Sutton and J.~Altarriba.
\newblock {Color associations to emotion and emotion-laden words: A collection
  of norms for stimulus construction and selection}.
\newblock \emph{Behav. Res.}, 48\penalty0 (2):\penalty0 686--728, Jun 2016.
\newblock ISSN 1554-3528.
\newblock \doi{10.3758/s13428-015-0598-8}.

\bibitem[Yamins et~al.(2014)Yamins, Hong, Cadieu, Solomon, Seibert, and
  DiCarlo]{Yamins2014}
D.~L.~K. Yamins, H.~Hong, C.~F. Cadieu, E.~A. Solomon, D.~Seibert, and J.~J.
  DiCarlo.
\newblock Performance-optimized hierarchical models predict neural responses in
  higher visual cortex.
\newblock \emph{Proceedings of the National Academy of Sciences}, 111\penalty0
  (23):\penalty0 8619--8624, May 2014.
\newblock \doi{10.1073/pnas.1403112111}.
\newblock URL \url{https://doi.org/10.1073/pnas.1403112111}.

\end{thebibliography}
\end{small}

\end{document}